\documentclass[10pt,twocolumn,letterpaper]{article}

\usepackage{cvpr}
\usepackage{times}
\usepackage{epsfig}
\usepackage{graphicx}
\usepackage{amsmath}
\usepackage{amssymb}

\usepackage{float}
\usepackage{epstopdf}
\usepackage{varioref}
\usepackage[utf8]{inputenc}
\usepackage{authblk}

\usepackage[pagebackref=true,breaklinks=true,letterpaper=true,colorlinks,bookmarks=false]{hyperref}

\cvprfinalcopy 

\def\cvprPaperID{****} 

\ifcvprfinal\fi
\begin{document}

\title{Image Captioning with Integrated Bottom-Up and Multi-level Residual Top-Down Attention for Game Scene Understanding}
\author{{Jian Zheng}$^{*1}$, Sudha Krishnamurthy$^{2}$, Ruxin Chen$^{2}$, Min-Hung Chen$^{3}$, Zhenhao Ge$^{2}$, Xiaohua Li$^{1}$\\
\textnormal{$^{1}$ State University of New York at Binghamton\\
$^{2}$ Sony Interactive Entertainment LLC\\
$^{3}$ Georgia Institute of Technology\\
$^{}$\{jzheng65, xli\}@binghamton.edu}, \{sudha.krishnamurthy, ruxin.chen, zhenhao.ge\}@sony.com, cmhungsteve@gatech.edu}
\date{}

\maketitle

\footnotetext[1]{This work is done during the first author's internship at SIE.}
\begin{abstract}
Image captioning has attracted considerable attention in recent years. However, little work has been done for game image captioning which has some unique characteristics and requirements. In this work we propose a novel game image captioning model which integrates bottom-up attention with a new multi-level residual top-down attention mechanism. Firstly, a lower-level residual top-down attention network is added to the Faster R-CNN based bottom-up attention network to address the problem that the latter may lose important spatial information when extracting regional features. Secondly, an upper-level residual top-down attention network is implemented in the caption generation network to better fuse the extracted regional features for subsequent caption prediction. We create two game datasets to evaluate the proposed model. Extensive experiments show that our proposed model outperforms existing baseline models.
\end{abstract}
\section{Introduction}

Game image captioning is to enable machines to comprehend game scenes and output a decent description for a given image frame of game videos. Game image captioning can help players to have a better understanding of the game scenes and, especially, can provide special assistance to visually-impaired players or machine players. Nevertheless, according to our knowledge, little work has been done for image captioning with game images.
There are no public datasets available for game image captioning. Game image captioning datesets are different from conventional image captioning datasets because game images involve more action related contents. Unlike conventional action recognition, game image captioning needs to generate more delicate and semantic sentences to describe the complex game scenes instead of just a few words of action description. Although game images may be generated artificially, the source codes and image descriptions are usually not available to captioning machines.

Attention has been widely studied in image captioning \cite{vinyals2015show}. Top-down attention alone is not enough to learn the attention distribution well, especially when the attention network is weakly-supervised. More recently, some efforts have been conducted to exploit both top-down attention and bottom-up attention
\cite{anderson2018bottom}, where bottom-up attention can be implemented based on Faster R-CNN \cite{ren2015faster}. However, existing bottom-up attention network may neglect some important spatial information in the detected salient regions. 

In this paper, we propose a new image captioning model for game scene understanding by integrating bottom-up attention with a novel multi-level residual top-down attention mechanism. Specifically, a lower-level residual top-down attention network is added into the Faster R-CNN based bottom-up attention network to exploit better spatial information. An upper-level residual top-down attention network is implemented in the caption generation network to selectively attend to certain regions for better caption prediction. 
In addition, we create two game image datasets to validate the performance of the proposed model.


\section{Related Work}
Image captioning has been a hot research topic during the past several years \cite{johnson2016densecap}\cite{rennie2017self}. Many top-down visual attention mechanisms have been developed and adopted in image captioning \cite{xu2015show}\cite{liu2017attention}\cite{lu2017knowing}\cite{anderson2018bottom}\cite{li2017image}. Recently, some approaches with both bottom-up attention and top-down attention are proposed \cite{jin2015aligning,pedersoli2016areas}. In particular, a combined bottom-up and top-down attention mechanism is proposed in \cite{anderson2018bottom}, which is the current state-of-the-art in image captioning model. This paper enhances \cite{anderson2018bottom} with a novel multi-level residual top-down attention mechanism and addresses the specific problem of game image captioning.

\begin{figure}[!tp]
	\begin{center}
		\includegraphics[width=1.0\linewidth]{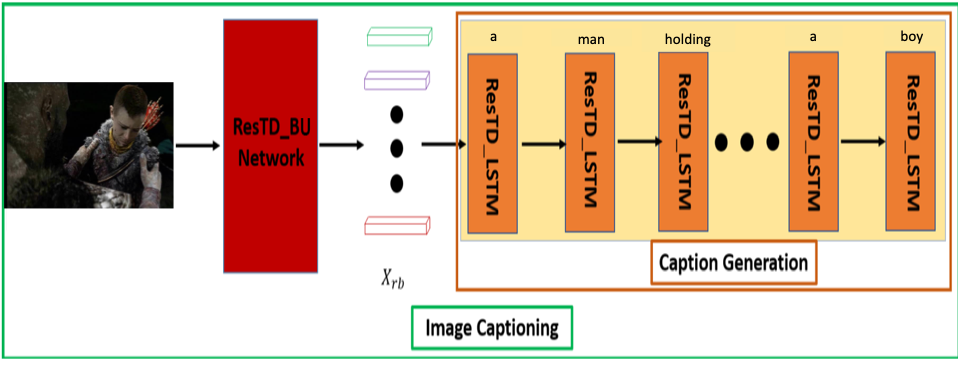}
	\end{center}
		\vspace{-0.05in}
	\caption{Block diagram of the proposed image captioning network.}
		\vspace{-0.05in}
	\label{fig:1}
	
\end{figure}

\section{Method}
Fig. \ref{fig:1} shows our proposed image captioning model. The model consists of mainly two parts: a Residual Top-Down Bottom-Up (ResTD\_BU) attention network and a Caption Generation network. ResTD\_BU extracts regional feature vectors from an image, while the Caption Generation network takes the extracted feature vectors of different salient regions as input and predicts the natural language caption word by word.



\subsection{ResTD\_BU Network}

\begin{figure}[!tp]
	\begin{center}
		\includegraphics[width=1.0\linewidth]{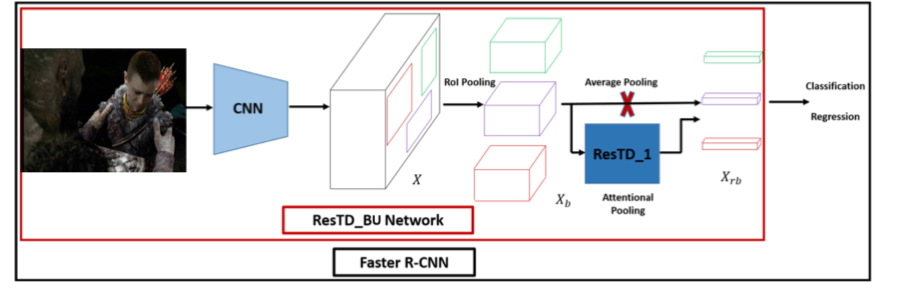}
	\end{center}
		\vspace{-0.05in}
	\caption{The ResTD\_BU Network: Faster R-CNN based bottom-up attention network with an integrated lower-level residual top-down attention network ResTD\_1.}
		\vspace{-0.05in}
	\label{fig:2}
\end{figure}

The ResTD\_BU network uses Faster R-CNN to realize bottom-up attention. Nevertheless, we replace the simple average pooling used in \cite{anderson2018bottom} with a novel residual network to introduce top-down attention, as illustrated with the red crossing and the ResTD\_1 block in Fig. \ref{fig:2}. 

Our ResTD\_BU network takes a given image as input and feeds it to a deep ConvNet (we use ResNet-101 \cite{he2016deep}) to learn the intermediate image feature representations. Then a Region Proposal Network (RPN) is applied over the intermediate feature maps to generate a large number of region proposals, among which only a small portion will be selected by applying non-maximum suppression (NMS) and intersection-over-union (IoU) threshold. The output of the RoI pooling layer is a set of $N$ region feature maps, described as ${X_b} = \{X_{b1}, X_{b1}, ..., X_{bN}\}$, where $X_{bi} \in \mathcal{R}^{n_1 \times n_2 \times D}$. We used $n_1 = n_2 = 7, D = 2048$ in our experiments. 

Next, instead of applying global average pooling, we propose to use the ResTD\_1 network to implement lower-level top-down attention on the region feature maps. Specifically, we take each ${X}_{bi}$ as input and feed it to ResTD\_1 to obtain a feature vector ${\bf x}_{rbi}$ for each region. 
In the end, we have a set of regional feature vectors ${X}_{rb} = \{{\bf x}_{rb1}, {\bf x}_{rb2}, ..., {\bf x}_{bN}\}$ for each given image.

\begin{figure}[!tp]
	\begin{center}
		\includegraphics[width=0.8\linewidth]{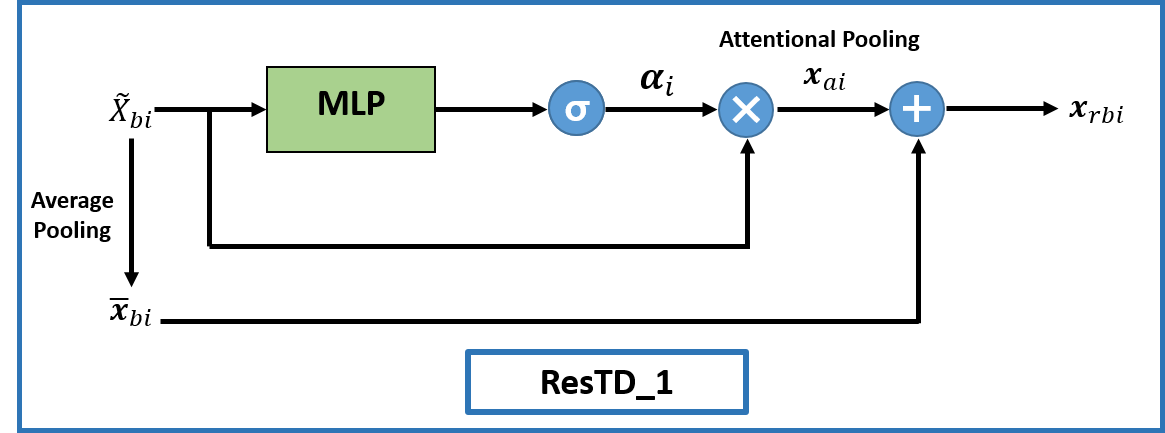}
	\end{center}
		\vspace{-0.05in}
	\caption{ResTD\_1: Lower-level residual top-down attention network.}
		\vspace{-0.05in}
	\label{fig:3}
\end{figure}

The ResTD\_1 network is implemented as in Fig. \ref{fig:3}, which is inspired by the residual network architecture \cite{he2016deep}. The MLP is a two-layer fully-connected network, $\sigma$ is the softmax function, $\times$ and $+$ are element-wise product and summation, respectively. The attention distribution among input data might not be learned well, thus important information could be lost during feature fusion. The averagely pooled feature vector $\bar{x}_{bi}$ from $\tilde{X}_{bi}$ can make up for the image information loss, where $\tilde{X}_{bi} \in \mathcal{R}^{(n_1 \times n_2) \times D}$.

\subsection{Caption Generation Network}

\begin{figure}[t]
	\begin{center}
		\includegraphics[width=1.0\linewidth]{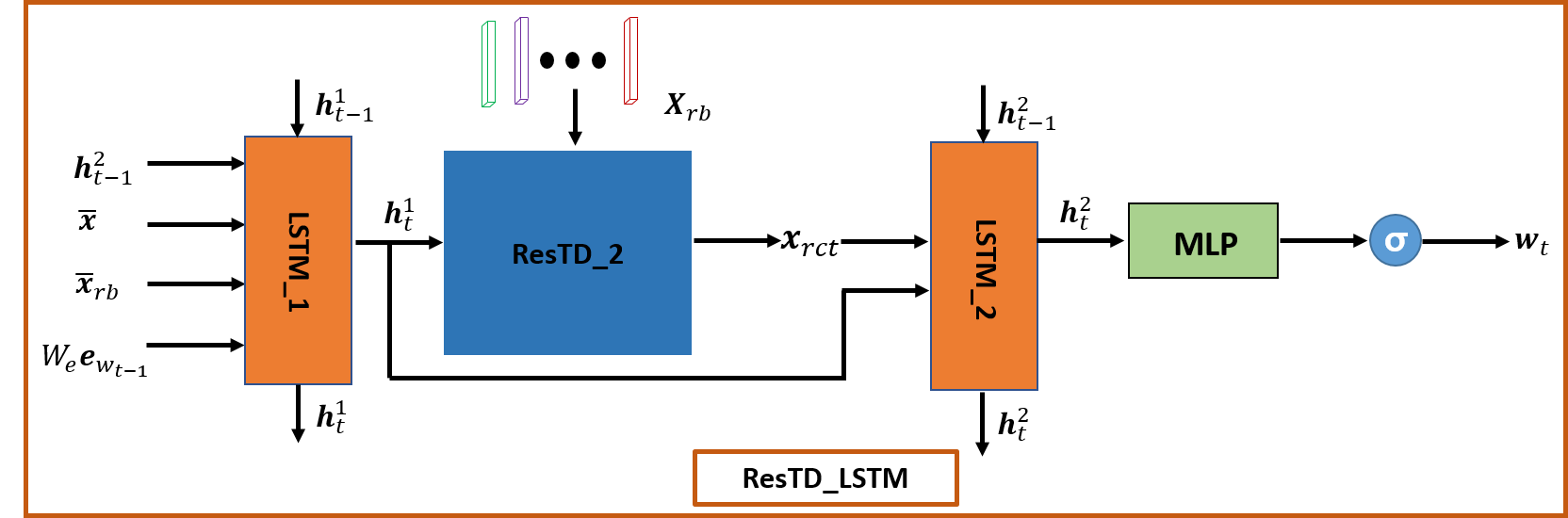}
	\end{center}
		\vspace{-0.05in}
	\caption{Block diagram of a ResTD\_LSTM unit.}
		\vspace{-0.05in}
	\label{fig:4}
\end{figure}

The Caption Generation network consists of a set of residual top-down attention based LSTM units, which we call ResTD\_LSTM. 
Top-down attention makes the network attend to selective regions when generating caption words. The network uses $X_{rb}$ as input. As shown in Fig. \ref{fig:4}, each ResTD\_LSTM unit has three components: an upper-level residual top-down attention network (RestTD\_2), an LSTM for contextual information embedding (LSTM\_1), and an LSTM for caption prediction (LSTM\_2).


LSTM\_1 is used to combine image feature information with context information. Image feature information includes the global bottom-up feature vector ${\bf {\bar x}}_{rb}$, which is obtained from $X_{rb}$ through average pooling, and the global feature vector of the entire image ${\bf {\bar x}}$, which is obtained through global average pooling over $X$. Context information consists of two components, i.e., $W_e e_{w_{t-1}}$, which is the word embedding predicted from the previous time step, and ${\bf h}^2_{t-1}$, which is the hidden state of LSTM\_2. The LSTM\_1 mixes all these information to get a general contextual vector ${\bf h}^1_{t}$. 

\begin{figure}[t]
	\begin{center}
		\includegraphics[width=0.8\linewidth]{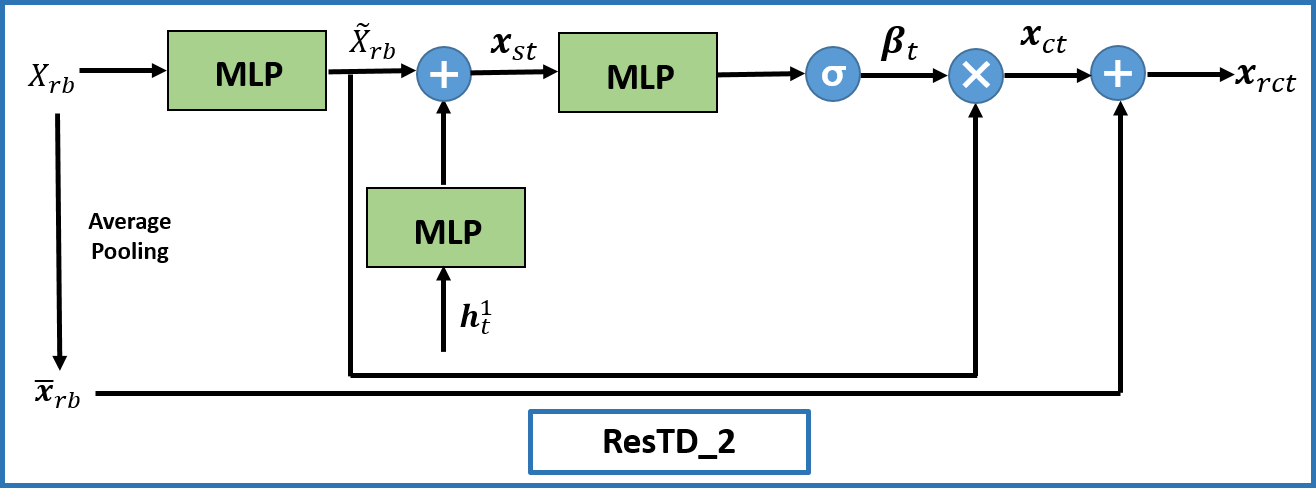}
	\end{center}
		\vspace{-0.05in}
	\caption{ResTD\_2: Upper-level residual top-down attention network.}
		\vspace{-0.05in}
	\label{fig:5}
\end{figure}

The vector ${{\bf h}^1_{t}}$, as well as the set of region feature vectors $X_{rb}$ are fed into ResTD\_2 to learn the attention distribution over the $N$ regions and to output the contextual vector ${\bf x}_{rct}$. The implementation of ResTD\_2 is shown in Fig. \ref{fig:5}. In addition to ${\bf x}_{rct}$, the input to the caption prediction cell LSTM\_2  includes the hidden state ${{\bf h}^2_{t-1}}$. To predict the next word ${\bf w}_t$, we feed the output of LSTM\_2 to a single-layer perceptron MLP, followed by a softmax function. 


\section{Datasets \& Experiments}

To train the ResTD\_BU network, we use Visual Genome \cite{krishna2017visual} and a game image dataset named GD\_Det, which is created by us for game object detection. To evaluate the overall image captioning model on game images, we use V-COCO \cite{gupta2015visual} and our own game image captioning dataset GD\_ImgCap. 

For game scene understanding, we are more interested in game objects and associated actions. Therefore, we only use the same 1600 object classes as \cite{anderson2018bottom} in Visual Genome. Our GD\_Det dataset contains 7 object classes: human, gun, axe, sword, monster, car, motorcycle. We split the dataset into 4,725 training images, 834 validation images, and 1,390 testing images. Since Visual Genome is much larger than GD\_Det, we train the ResTD\_BU network with these two datasets separately. 
We implement the ResTD\_BU network based on \cite{chen2017implementation}. 

We created our image captioning dataset GD\_ImgCap with 7,558 game images collected from 131 different game videos, including 5,920 training images, 658 validation images, as well as 606 test images. The length of game image captions ranges from 2 to 19. On average, there are 2 captions for each game image. 
We combine V-COCO with GD\_ImgCap as the final game image captioning dataset. As a result, there are 15,331 training images, 1693 validation images, and 606 test images. We implement the image captioning model based on \cite{Luo2017}. 

We compare the performance of our proposed image captioning model with four existing models, i.e., SAT \cite{xu2015show}, TD, Att2in2 \cite{rennie2017self}, and AdaAtt \cite{lu2017knowing}. TD is a baseline image captioning model with an architecture similar to our model. The difference is that in TD, we remove ${\bf {\bar x}}$ from LSTM\_1 and remove ${\bf {\bar x}}_{rb}$ from ResTD\_2. The inputs to these four models are our baseline bottom-up attention regional features $X_{rb}$ extracted without lower-level residual top-down attention introduced.

\section{Experiment Results}
{\bf Detection Results:}
Experiment results shown in Table \ref{Table:1} demonstrate the effectiveness of the proposed lower-level residual top-down attention network. Regular top-down attention ({\bf Att}) improves the detection performance of Faster R-CNN, but our residual top-down attention network ({\bf ResAtt}) has the best performance. 

\begin{table}[t]
\small
	\begin{center}
		\begin{tabular}{c c c}
			\hline
			 & GD\_Det  & Visual Genome\\
			\hline
			No\_Att & 58.22\% & 4.62\% \\
			Att & 60.15\% & 6.36\%  \\
			ResAtt & {\bf 61.88\%} & {\bf 7.34\%}  \\
			\hline
		\end{tabular}
	\end{center}
		\vspace{-0.05in}
	\caption{Performance (mAP) comparison of Faster R-CNN object detection with/without top-down attention.}
		\vspace{-0.05in}
	\label{Table:1}
\end{table} 

{\bf Captioning Results:}
We report the captioning results of our proposed model ({\bf Ours}) with the other four models on GD\_ImgCap in Table \ref{Table:2}. It can be seen  that our model outperforms all the other four models. Especially, by comparing our model with the baseline model TD, we come to the conclusion that integrating bottom-up and multi-level residual top-down attention drastically improves captioning performance. 

\begin{table}[t]
\small
	\begin{center}
		\begin{tabular}{c c c c c c}
			\hline
			 & avg\_BLEU  & CIDEr & METEOR & ROUGE & SPICE\\
			\hline
			SAT & 16.53 & 99.37 & 13.32 & 33.74 & 26.97\\
			TD & 17.89 & 117.77 & 13.82 & 35.83 &  30.53\\
			Att2in2 & 18.74 & 118.40 & 14.21 & 36.98 & 30.17 \\
			AdaAtt & 19.18  & 127.84 & 14.43 & 36.66 & 30.09 \\
			Ours & {\bf 19.59} & {\bf 127.90} & {\bf 15.78} & {\bf 38.67}  & {\bf 32.67} \\
			\hline
		\end{tabular}
	\end{center}
		\vspace{-0.05in}
	\caption{Performance comparison of our game captioning model with competing existing models on GD\_ImgCap.}
		\vspace{-0.05in}
	\label{Table:2}
\end{table}

\begin{table}[t]
   \small
	\begin{center}
	\setlength{\tabcolsep}{1.5pt}
		\begin{tabular}{l c c c c c}
			\hline
			 & avg\_BLEU & CIDEr & METEOR & ROUGE & SPICE\\
			\hline
			BU\_Only & 18.28 & 115.78 & 14.43 & 37.08 & 30.07\\
		    BU+Td & 18.13 & 115.80 & 15.14 & 37.20 & 30.79 \\
			BU+ResTd & {\bf 19.59} & {\bf 127.90} & {\bf 15.78} & {\bf 38.67} & {\bf 32.67} \\
			\hline
		\end{tabular}
	\end{center}
	\vspace{-0.05in}
	\caption{The captioning performance under various model variants.}
	\vspace{-0.05in}
	\label{Table:4}
\end{table}


{\bf Ablation Study:}
To better understand our residual top-don attention, we compared our proposed model with its two variants: {\bf BU\_only} where no top-down attention was used, and {\bf BU+Td} where the residual top-down attention was replaced by conventional top-down attention.  Results shown in Table \ref{Table:4} indicates that  residual top-down attention greatly improves game captioning performance. 
our proposed model outperformed conventional top-down attention model, and game captioning performance was improved greatly by introducing the lower-level residual top-down attention. 


\begin{figure}[t]
	\begin{center}
		\includegraphics[width=1.0\linewidth]{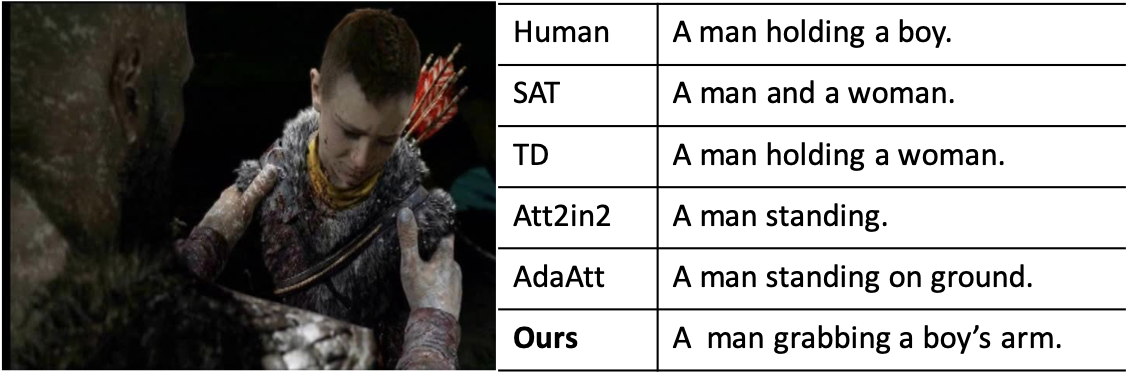}
	\end{center}
	\vspace{-0.05in}
	\caption{Captions generated from our model and some existing models for a game image.}
	\label{fig:10}
	\vspace{-0.05in}
\end{figure}

{\bf Caption Example:}
Fig. \ref{fig:10} shows the captions generated from a game image. It can be seen that our proposed model outperforms the other four models in the sense that it generates more detailed caption that is closer to the human-annotated ground truth caption. 

\section{Conclusion}
In this paper, we propose a novel game image captioning model with integrated bottom-up and multi-level residual top-down attention. To extract better regional feature representations, we introduce a lower-level residual top-down attention network into the Faster R-CNN based bottom-up network. In order to attend better to different regions during caption prediction, we employ an upper-level residual top-down attention network in the caption generation network. Two game image datasets are created. Extensive experiments are conducted to demonstrate the effectiveness of the proposed model on game images.

{\small
\bibliographystyle{ieee}
\bibliography{egbib}
}

\end{document}